\documentclass[twoside,11pt]{article}

\usepackage{jmlr2e}
\usepackage{amsmath}
\usepackage{subfig}
\usepackage{amsmath}
\usepackage{natbib}

\ShortHeadings{Automatic Seizure Detection}{Thodoroff and Pineau}
\firstpageno{1}

\begin{document}

\title{Learning Robust Features using Deep Learning for Automatic Seizure Detection}

\author{\name Pierre Thodoroff \email pierre.thodoroff@mail.mcgill.ca \\
       \addr Reasoning and Learning Lab\\
       School of Computer Science, McGill University\\
       Montreal, Canada\\
       \name Joelle Pineau \email jpineau@cs.mcgill.ca \\
       \addr Reasoning and Learning Lab\\
       School of Computer Science, McGill University\\
       Montreal, Canada\\
        \name Andrew Lim \email andrew.lim@utoronto.ca \\
       \addr Sunnybrook Health Sciences Centre\\
       University of Toronto\\
       Toronto, Canada\
}

\maketitle

\begin{abstract}
We present and evaluate the capacity of a deep neural network to learn robust features from EEG to automatically detect seizures. This is a challenging problem because seizure manifestations on EEG are extremely variable both inter- and intra-patient. By simultaneously capturing spectral, temporal and spatial information our recurrent convolutional neural network learns a general spatially invariant representation of a seizure. The proposed approach exceeds significantly previous results obtained on cross-patient classifiers both in terms of sensitivity and false positive rate. Furthermore, our model proves to be robust to missing channel and variable electrode montage.
\end{abstract}
\section{Introduction}
Epilepsy is a neurological disorder affecting more than 50 million people worldwide~\citep{epilepsy_world}. Monitoring of brain activity through electroencephalograms (EEG) is the standard technique for the diagnosis of epilepsy. In current clinical practice, EEG readings must be analyzed by trained neurologists to identify characteristic patterns of the disease, such as seizures and pre-ictal spikes. However this visual analysis is extremely laborious, taking several hours to analyze one day of recording from a single patient, and it requires scarce highly trained professionals. In many clinics, this limits how many patient recordings can be analyzed. Furthermore, it is not always feasible to have access to neurologists especially in developing countries (e.g. see The Bhutan Epilepsy Project\footnote{\url{http://www.bhutanbrain.com}}). Those limitations have motivated efforts to develop automated approaches to seizure detection.

The problem of automatic seizure detection has been extensively studied.  Most work to date uses expert hand-crafted features characteristic of seizure manifestations in EEG; many proposed methods rely on spectral information~\citep{review_seizure}, whereas some of them capture the temporal aspect of a seizure~\citep{shoeb}.  However it is well known that epileptic seizures are highly non-stationary phenomena and seizure manifestations in EEG are extremely variable both within a patient over time, and between different patients~\citep{clinical_seizure}. 

In an effort to improve the generalization error in automated seizure detection both intra- and inter-patient, we study the potential of deep learning in a supervised learning framework to automatically learn more robust features. Indeed, features designed by deep learning models have proven to be more robust than hand-crafted features in various field, including computer vision and speech recognition~\citep{lecun-bengio-95a}. The architecture proposed in our work consists of a recurrent convolutional neural network. It is designed to simultaneously capture spectral, temporal and spatial information while learning a general spatially-invariant representation of a seizure (particularly relevant for cross-patient classifier).

We apply the method to a large publicly available data set, the Children's Hospital of Boston-Massachusetts Institute of Technology dataset (CHB-MIT).  Results show that our approach can match state-of-the-art performance in terms of sensitivity and false positive rate on patient-specific seizure detection. Moreover, results with this model on the cross-patient seizure detection task exceed previous results by a significant margin.  Finally, we show that the model is robust to missing channels and different electrode montage, thus making it practical for realistic clinical settings.

\section{Problem Definition}
Epileptic seizures are characterized by episodes of excessive or abnormal synchronous neuronal activity in the brain.  Seizures can be accompanied by clinical neurological symptoms, such as loss of, or alterations, in consciousness, abnormal movements, or abnormal sensory phenomena, and are therefore associated with considerable neurological morbidity.  Broadly speaking, seizures can be anatomically classified into two categories: those of partial onset, which arise from a specific brain region, with or without secondary generalization; and those of generalized onset, which arise synchronously from the brain as a whole.  Seizures can vary dramatically between patients, and even within individual patients.

There are two phases to the treatment of seizures. In the acute phase, medications can be administered to abort an ongoing seizure. In the chronic phase, medications are taken on a daily basis to prevent further seizures. In the case of focal seizures, surgical resection of the region or regions of the brain generating the seizures can be used to prevent further seizures.  All of these treatments require accurate detection and classification of seizures as either partial or generalized onset. Indeed surgical management of partial onset seizures requires identification of the specific region of the brain generating the seizures. Seizure detection is also used to monitor patients under treatment or surgical resection to assess the efficacy of the procedures undertaken.

The primary diagnostic tool for detecting seizures is electroencephalography (EEG): the continuous measurement of brain-generated electrical potentials by means of electrodes placed on the scalp, or directly on the surface of the brain.  Continuous EEG recordings obtained for the purposes of recording seizures are typically of hours to days duration. They are visually analyzed by trained neurologists to detect seizures, classify them, and, if applicable, identify where in the brain they are originating from. 
As mentioned earlier, this visual analysis of EEG is laborious and costly, motivating the development of software to perform automatic seizure detection.  Indeed, specialized detectors can be used to enhance monitoring of patients under treatment or post surgical resection. Whereas more general detectors can enhance diagnosis and treatment planning on new patient. This is particularly relevant in developing countries where access to a knowledgeable expert is not possible.

Depending on the clinical applications, two situations arise. If previously annotated patient data is available, one can design a patient-specific detector. Otherwise, one needs a model that is able to detect seizures without patient-specific training data. Patient-specific detectors can be used to monitor patients under a particular treatment, whereas cross-patient detectors can be used to diagnosis a new patient and help plan potential treatment.

In this study, we evaluate our methods on the Children's Hospital of Boston-Massachusetts Institute of Technology dataset (CHB-MIT).  This is the biggest freely available dataset existing~\citep{shoeb}. Extensive work has been done on this dataset facilitating the comparison of methods~\citep{review_seizure,shoeb}. The CHB-MIT dataset contains 23 patients divided among 24 cases (a patient has 2 recordings, 1.5 years apart). The dataset consists of 969 Hours of scalp EEG recordings with 173 seizures. There exist various types of seizures in the dataset (clonic, atonic, tonic). The diversity of patients (Male, Female, 10-22 years old) and different types of seizures contained in the datasets are ideal for assessing the performance of our methods in realistic settings. In this paper, for patient-specific and cross-patient detection, the goal is to detect whether a 30 second segment of signal contains a seizure or not, as annotated in the dataset.

\section{Previous Work}

Two problems have been extensively studied in the past 35 years~\citep{gotman1}: online seizure prediction and offline seizure detection. Online predictions attempt to predict in advance when seizures will occur, with the possible goal of applying just-in-time treatment or intervention options~\citep{mormann07}.  In contrast, offline detection focuses on analyzing recordings that are completed and aims to automatically label these recordings for the purposes of diagnosis, monitoring or treatment planning~\citep{offline_eeg}. The main use of offline detectors is to replace the need for laborious visual analysis of day-long recordings. This paper focuses on the latter for its potential for clinical application.

We characterize the performance of models using sensitivity and false detection rate, as is standard in the seizure detection community~\citep{review_seizure}. Sensitivity measures the proportion of real seizures that were correctly identified by a classifier while the false detection rate indicates the number of false alarms raised by a detector per hour of recording. A balance between these 2 metrics must be attained. Some research papers report the specificity rather than false detection rate. However, it is important to note that a seemingly high specificity of 95\% is equivalent to a false positive rate of ~5/hours, assuming 30 seconds window (which is considered a poor performance).

\subsection{Patient specific detectors}
Recent research in automated seizure detection started using machine learning for patient-specific detectors. By using hand-crafted EEG features many publications designed accurate patient-specific detectors.
\cite{shoeb} achieved a sensitivity of 96\% and a low false positive rate of 0.08/hours using an SVM classifier over a combination of spectral, spatial, and temporal handcrafted features.  The results were published with the associated dataset (CHB-MIT) and can be considered a good benchmark. Similar results have been obtained on this dataset in various publications~\citep{chbmit_pat}.

\subsection{Cross patient detectors}
Cross patient detectors, in contrast, have proven to be much more challenging. Seizure manifestations in EEG can vary dramatically inter-patient (locations in the brain, shapes, durations), thus complicating the design of generalized seizure detectors. A recent large study by~\cite{multi} involving 205 patients  from 3 different epilepsy centers represents a good benchmark for the accuracy of current methods. Indeed, the inclusion of different data center reduces the bias of the method. They achieved an average sensitivity of 67\% and a false detection rate of 0.32~\citep{multi} on the CHB-MIT. However, it is interesting to note that their methods achieved an average sensitivity of 81\%  with a false detection rate of 0.29/hours on four other datasets. The only commercialized software to automatically detect seizures across patients only detected 61\% of the seizures with a false detection rate of 1.375/hour on the CHB-MIT dataset~\citep{reveal_cross}.

\section{Methods}

We propose a recurrent convolutional architecture designed to capture spectral, temporal and spatial patterns representing a seizure. Combined with an image-based representation of EEG incorporating domain knowledge, the model learns a patient-independent representation seizures. Figure 1 illustrates the pipeline of our detection architecture.  First, the multi-channel EEG signal is projected into an image representation using the method described in Sec.~\ref{sec:image}.  Second, a recurrent convolutional neural network is trained to predict whether or not the corresponding image contains a seizure.

\begin{figure}[htbp]
  \centering 
  \includegraphics[width=5in,height=3in]{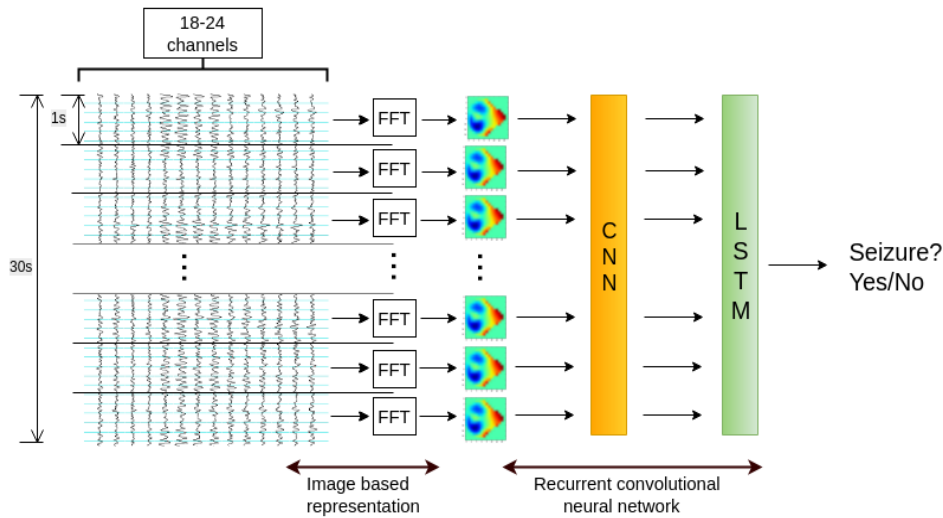} 
  \caption{Recurrent convolutional neural network using image-based representation of EEG.}
  \label{fig:full_model} 
\end{figure}

\subsection{Image based representation}
\label{sec:image}
To exploit the spatial locality existing in seizures we first create an image-based representation of EEGs, integrating spatial domain knowledge (electrodes montage), using a method designed by~\cite{bashivan}. The first step consists of  projecting the 3D coordinates of the patient electrodes onto a 2D surface. In order to preserve the distance between electrodes in the 3D plane, we project using Polar Projection~\citep{polar_proj}. Then, we assign to each electrode projection values in 3 channels representing the magnitude of different frequency bands (0-7,7-14,14-49 Hertz) in the given 1 second segment of the signal. Finally, to create a continuous image, we interpolate the values of each electrode projection using cubic interpolation. This creates images of shape (3x16x16). Each image has 3 color channel (1 for each frequency band) with height and width of 16 pixels.

\begin{figure}[htbp]
  \centering 
  \includegraphics[width=5in,height=2in]{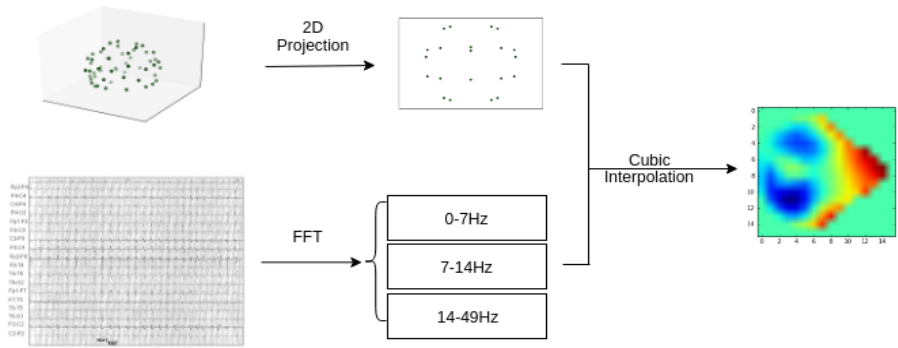} 
  \caption{Image-based representation of EEG }
  \label{fig:im_rep} 
\end{figure} 

\subsection{Recurrent convolutional neural network}

Convolutional neural networks are artificial neural networks inspired by the human visual cortex. They have been shown to achieve good results on challenging computer vision tasks~\citep{NIPS2012_4824}. In particular, their ability to extract representations that are robust to spatial translation~\citep{lecun-bengio-95a} makes them an promising candidate to detect seizures across different areas of the brain. In our setting, the convolutional layers are effectively learning a general spatially-invariant representation of a seizure.

\begin{figure}[htbp]
  \centering 
  \includegraphics[width=6in]{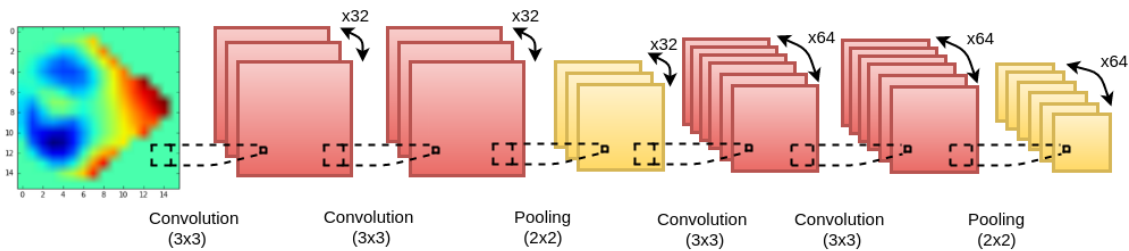} 
  \caption{Convolutional layers architecture }
  \label{fig:conv_model}
\end{figure}

A convolutional neural network consists of convolutional and sub-sampling layers, followed by a fully connected layer. The convolutional architecture used in this paper is shown in Figure~\ref{fig:conv_model}. It is inspired by a model that achieved state of the art on the ImageNet competition~\citep{NIPS2012_4824}. We use this neural architecture as a feature extractor, thus replacing the need for complex feature engineering that was used in previous work on seizure detection.

\begin{figure}[htbp]
  \centering 
  \includegraphics[width=3.8in,height=2.2in]{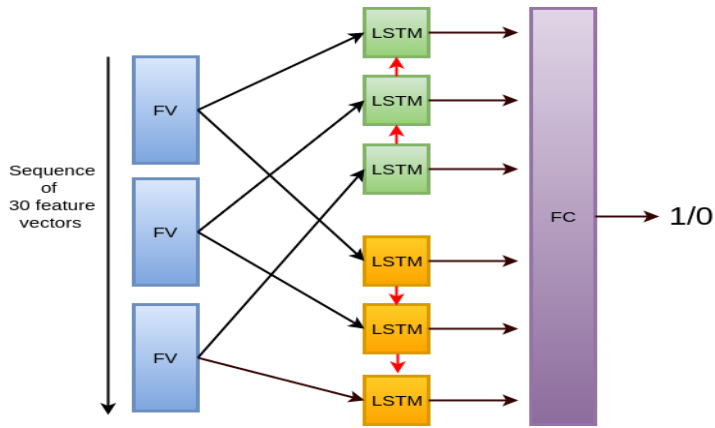} 
  \caption{Recurrent Neural Network architecture.  FV = Feature vector (64 dimension), LSTM = Long Short Term Recurrent node(128 hidden units), FC = Fully connected layer(512 hidden units)}
  \label{fig:recurrentmodel} 
\end{figure}

Recurrent neural networks, illustrated in Figure~\ref{fig:recurrentmodel}, are a special type of artificial neural network that contain loops in the neural architecture to allow for information to persist over time. More specifically, we use Long short Term Memory units (LSTM), which have proven effective to diffuse information over time sequences~\citep{Hochreiter:1997:LSM:1246443.1246450}. This is particularly relevant to analyze EEG data, given that seizures typically span several consecutive 1-second windows.  Bidirectional recurrent neural networks~\citep{graves2005framewise} exploit the fact that the output at time T depends on the previous elements but also on future ones. This is pertinent for automatic seizure detection. Indeed, in order to classify a specific window, neurologist often look at past and future windows. As we can see in Figure~\ref{fig:recurrentmodel}, the yellow LSTM units process the sequence in chronological order, whereas the green units process the sequence in reverse chronological order. 

To connect the convolutional network with the Recurrent network, we feed the output of the network shown in Figure~\ref{fig:conv_model} (where each image, built from 1s of data, produces an output vector of dimension 64) into one of the FV blocks of the architecture in Figure~\ref{fig:recurrentmodel}.  The recurrent architecture takes as input 30 such blocks (= 30 secs of EEG recording).

The concatenated recurrent convolutional architecture is trained jointly using gradient descent.  There are several hyper-parameters to select in such a model (see~\citep{Goodfellow-et-al-2016-Book} for more details). We sampled uniformly at random over the hyper-parameter space to optimize the parameters of the model. The search yielded the following parameters: Batch size = 128, Optimizer = rmsprop, Dropout = False, Learning rate = 0.001. Furthermore, using early-stopping~\citep{early_stopping}, we interrupt training as soon as the validation accuracy decrease (over-fitting).

For patient-specific detection, we train our neural model using only the patient's own data.  Due to the limited number of positive samples (seizures are typically relatively rare events), in our results below we test our model using a leave-one-out scheme. We test the accuracy of the model by training on N-1 seizures and testing on the withheld seizures. We repeat this process N times such that each seizure record is tested.

For cross-patient detection, we train our neural model using data for N-1 other patients and then test on the withheld patient. For the results below we repeat this process N times such that each patient is tested.

\subsection{Training neural network in a sample-efficient manner}

Deep learning models are powerful but sensitive to parametrization and training. Furthermore, seizures datasets suffer from severe class imbalance and few positive samples making the training precarious. Several methods have been proposed in  recent years to alleviate those problems.

First, by randomly subsampling the negative samples of the dataset we can re-balance the ratio between non-seizure and seizure data (from 1000/1 to 80/20) and thus facilitate the training. Indeed, deep learning hasn't proven to be very effective to handle unbalanced dataset~\citep{imbalance}.  Note that we only subsample the majority class during training. To test the accuracy of the model we use all the test data available in order to avoid over-optimistic results. It is important to multiply the prediction probability distribution by the appropriate constant to re-establish the training class distribution.

A second challenge is the overall lack of data for each patient.  Deep learning models are usually trained on millions of samples. In our case only 183 positive samples were available. We use pre-training~\citep{unsupervised} to help finding a good optimum as follows. We first train the convolutional layers alone to correctly classify 1 second windows. Then we train the entire model on sequences of 30 seconds using the convolutional weights learned previously as initialization weights. For patient-specific detectors, the amount of data available is even smaller (average of 8 seizures per patient). Using transfer learning, we first learn a general representation of a seizure on other patients and train the model to the specific patient using the weights previously learned as initialization.

Finally, in order to reduce the variance of the prediction probability distribution~\citep{zhou2002ensembling}, we apply an ensemble method by averaging over the predictions of three structurally identical model with different weight initializations.

\section{Results on the CHB-MIT dataset}

To evaluate the performance of our neural model we compare our patient-specific detector to the~\cite{shoeb} detector. For the cross-patient case we benchmark our model against the commercialized algorithm REVEAL~\citep{reveal_cross} using results from REVEAL CHB-MIT dataset that were published in~\cite{shoeb}'s PhD thesis.

\subsection{Patient specific detection}

Benchmarking is a complicated problem in seizure detection due to the different settings that research papers use and the disagreement existing across experts on the definition of a seizure (Ronnera et al., 2009). Especially for patient-specific detectors, it is complicated to compare algorithms that are in the 95-100\% sensitivity range. 

\begin{figure}%
    \centering
    \subfloat[Sensitivity of the patient specific detector designed in \cite{shoeb} thesis]{{\includegraphics[width=7cm,height=4cm]{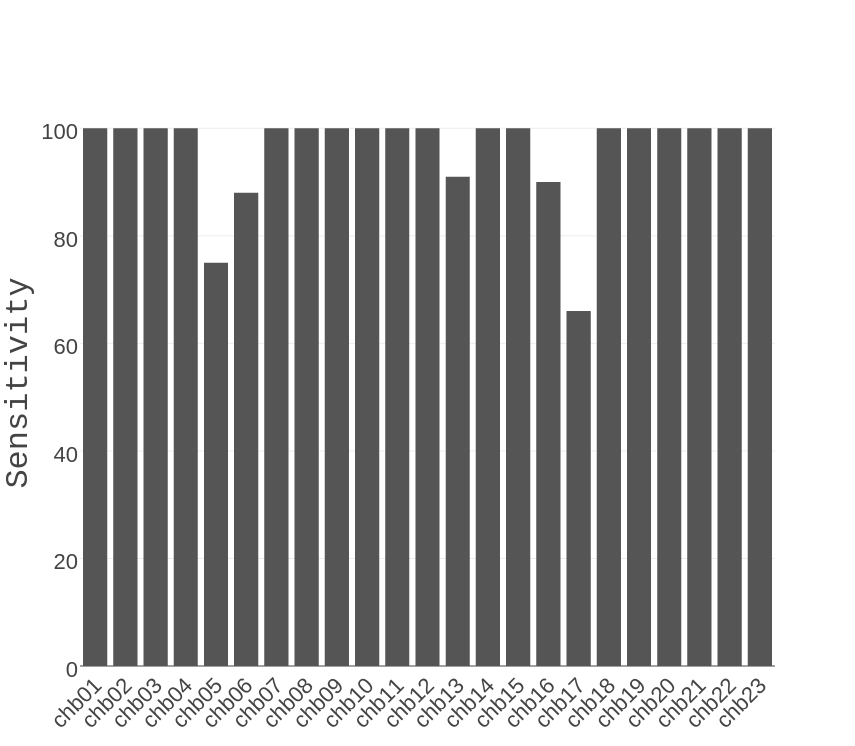} }}%
    \qquad
    \subfloat[False positive rate of the patient specific detector designed in \cite{shoeb} thesis]{{\includegraphics[width=7cm,height=4cm]{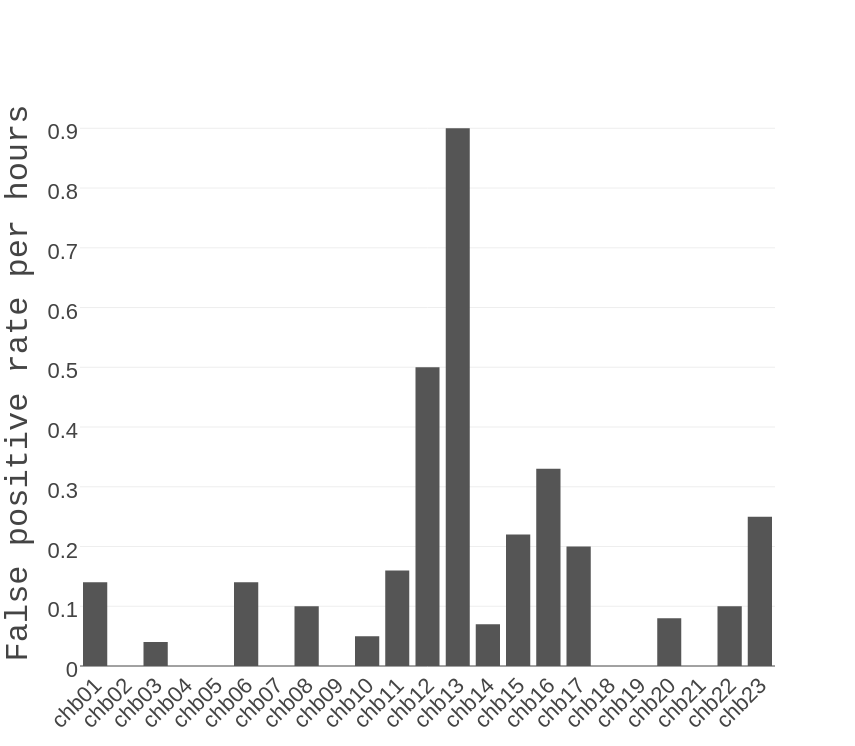} }}%
    \qquad
    \subfloat[Sensitivity difference (recurrent convolutional neural network minus Shoeb detector)  ]{{\includegraphics[width=7cm,height=4cm]{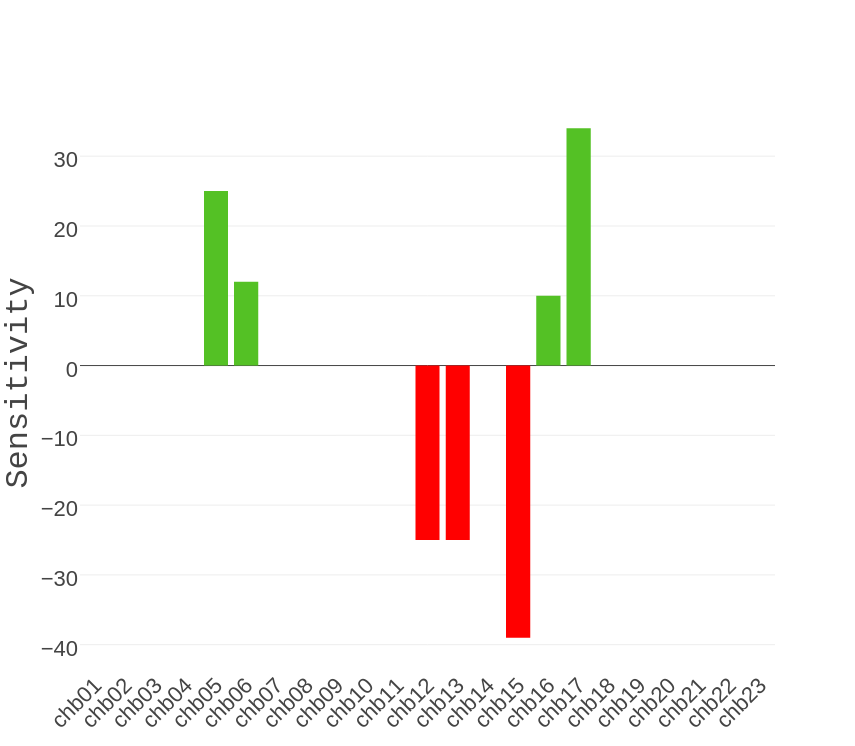} }}%
    \qquad
    \subfloat[False positive rate difference(Shoeb detector minus recurrent convolutional neural network)]{{\includegraphics[width=7cm,height=4cm]{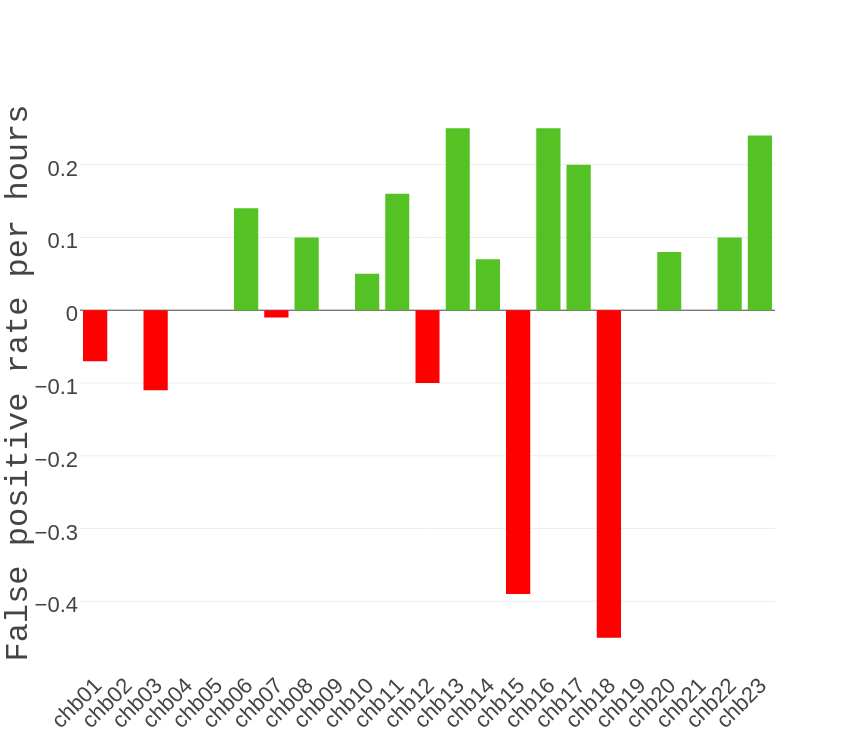} }}%
    \qquad
    \caption{Comparison of our patient specific model and \cite{shoeb} detector using hand crafted expert features and SVM}%
    \label{fig:patient-specific}%
\end{figure}

\begin{figure}[htbp]
  \centering 
  \includegraphics[width=5in]{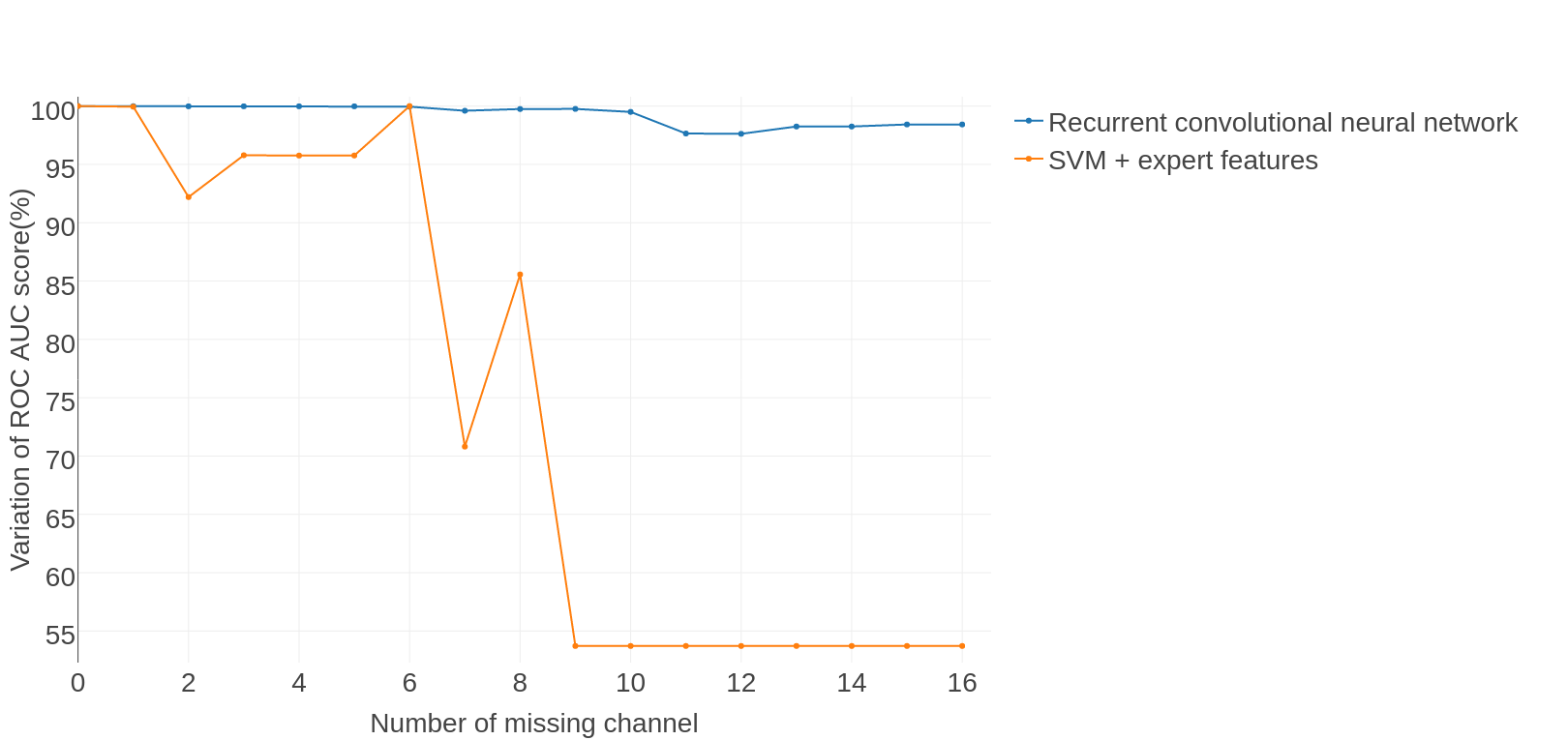} 
  \caption{Robustness to missing channel of Recurrent convolutional network vs SVM with expert features. The SVM model is our re-implementation of the approach proposed in~\citep{shoeb}.}
  \label{fig:roc} 
\end{figure}

Figure~\ref{fig:patient-specific} compares sensitivity and false positive rates achieved by Shoeb's SVM detector with our proposed neural architecture.  Overall both methods seem to obtain similar results for, both sensitivity and false positive rate, patient-specific detectors. Both methods achieve an accuracy that would allow the detector to be used in clinical settings for patient-specific classifier\citep{clinical_bench}. However, our neural model turns out to be significantly more robust to missing channel as observed in Figure~\ref{fig:roc}.

\subsection{Cross patient detection}
Traditional methods typically do not perform very well on new patients, due to their low capacity to generalize well across different seizure patterns. Indeed, REVEAL sensitivity drops significantly on many of the patients compare to patient specific detectors. 
The main advantage of our deep architecture lies in its ability to generalize well. 

\begin{figure}%
    \centering
    \subfloat[Sensitivity of the cross patient algorithm REVEAL~\citep{reveal_cross} ]{{\includegraphics[width=7cm,height=4cm]{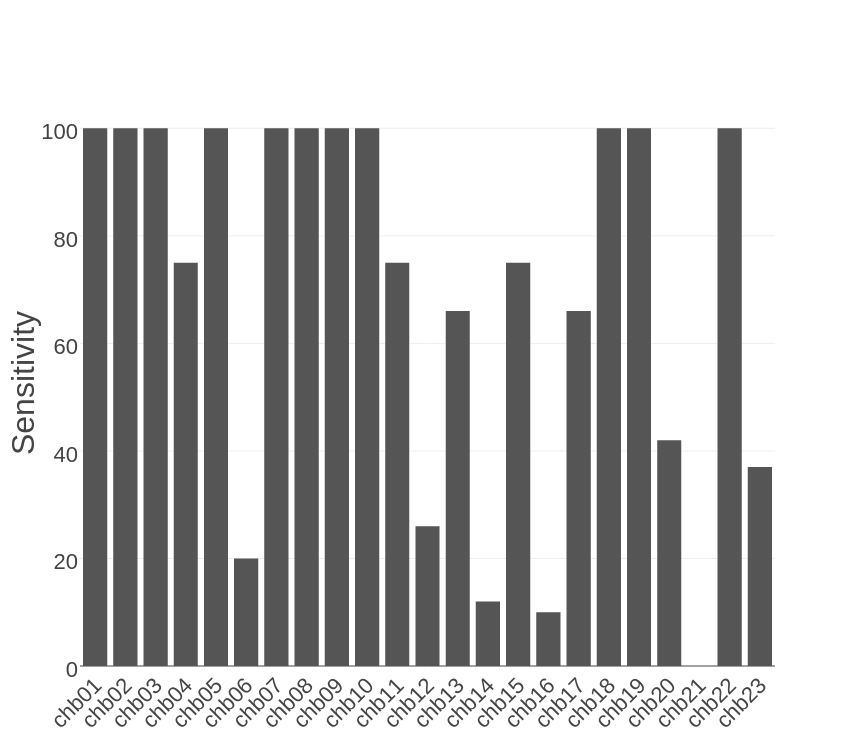} }}%
    \qquad
    \subfloat[False positive rate of the cross patient algorithm REVEAL~\citep{reveal_cross}]{{\includegraphics[width=7cm,height=4cm]{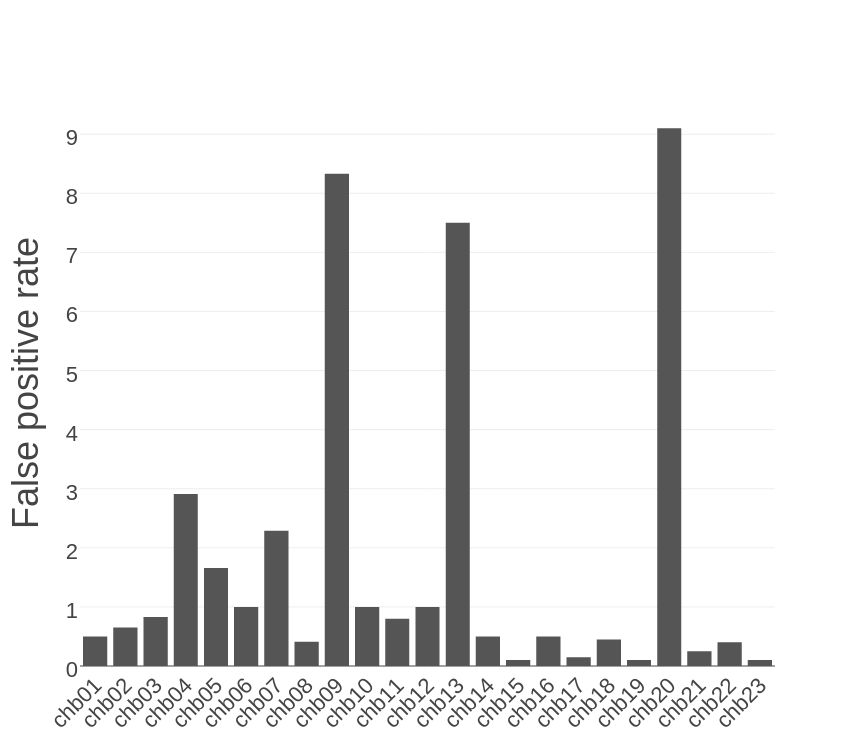} }}%
    \qquad
    \subfloat[Sensitivity difference (recurrent convolutional neural network minus REVEAL)  ]{{\includegraphics[width=7cm,height=4cm]{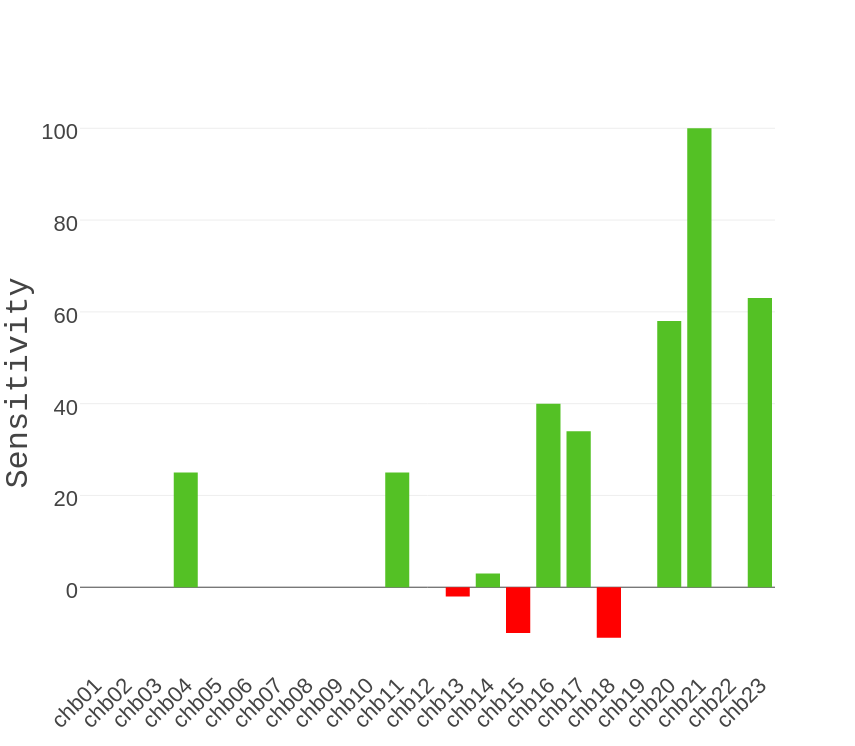} }}%
    \qquad
    \subfloat[False positive rate difference(REVEAL minus recurrent convolutional neural network)]{{\includegraphics[width=7cm,height=4cm]{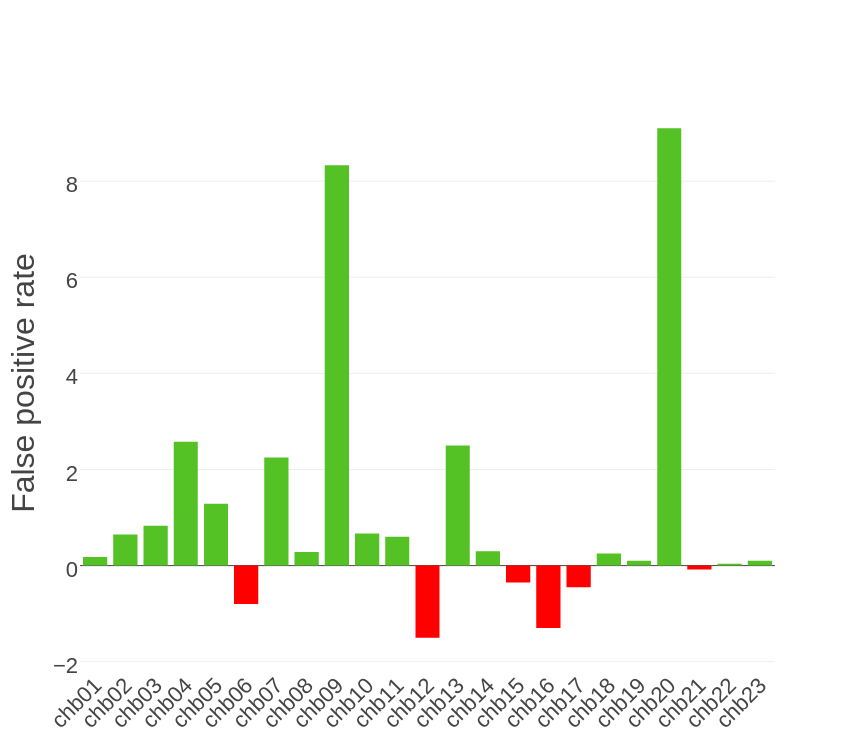} }}%
    \qquad
    
    \caption{Comparison of our cross patient specific model and the REVEAL algorithm from \cite{reveal_cross} applied to the CHB-MIT dataset by \cite{shoeb}}%
    \label{fig:cross-patient}%
\end{figure}

We can observe from Figure~\ref{fig:cross-patient} that our average sensitivity (85\%) is significantly higher than the average sensitivity obtained by REVEAL (67\%). Furthermore, a significant decrease of the false positive rate is achieved (from 1.7/hours to 0.8/hours).

\section{Discussion}

In this paper we proposed a new neural model for seizure detection that automatically learns robust features from spatial, temporal and frequency information contained in the EEG signal.
We observe that our model reaches state-of-the-art performance on patient specific detectors, furthermore its ability to learn a general representation of a seizure leads to significant improvement in cross-patient detection performance.  Automation of this process can enhance the diagnosis, monitoring, and treatment planning for patients with epilepsy.  The technology could be particularly useful in developing country where access to neurologist is impossible.

\begin{figure}[htbp]
  \centering 
  \includegraphics[width=3in,height=1.3in]{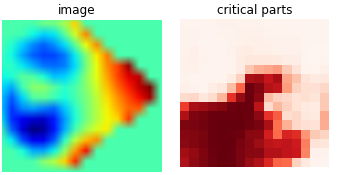} 
  \caption{Occlusion plot}
  \label{fig:Occlusion} 
\end{figure}

Our results also show that the image-based representation has clinical advantage. Indeed, the interpolation methods used between electrodes allows for classification of EEG with different electrode montage (several patient in CHB-MIT). 
Another advantage of this architecture lies in its ability to detect where a seizure is happening in the brain. Indeed, by occluding part of the image and testing the model's ability to predict the correct label we can define which area of the brain is responsible for the activation. Using a sliding window, we successively occlude part of the image and attempt to classify the occluded image correctly. If we fail to classify correctly the image, it means the area occluded is critical. We can see from Figure~\ref{fig:Occlusion} that the seizure is happening around the left side of the parietal, frontal or temporal lobes. One of the reasons neurologists analyze EEG is to locate from which part of the brain the seizures are coming from.

While achieving good results on cross-patient trials, the sensitivity for four of the patients was  low. The pattern of their seizures was probably different from what was available in the training set, highlighting the need for more data. Moreover, the false positive rate on cross-patient detectors attained is higher than for patient-specific detection. 

All the methods defined in section 4.3 considerably helped train our neural network in a sample-efficient manner. However, the variance of the prediction distribution remains high. Indeed by training on a small subset of the negative samples and testing on a significantly larger set of negative sample,  small parameter change can impact considerably the false positive rate. This is an inherent problem of deep learning architecture applied to small dataset. Future work on unsupervised methods to pre-train the network could address this issue.

\section{Acknowledgements}
The authors wish to thank Edith Law and Evgeny Naumov for
helpful discussions of this work.  Financial support was provided by NSERC
and CIHR via the Collaborative Health Research Projects program.
\nocite{chollet2015keras}
\bibliography{paper}

\begin{thebibliography}{23}
\providecommand{\natexlab}[1]{#1}
\providecommand{\url}[1]{\texttt{#1}}
\expandafter\ifx\csname urlstyle\endcsname\relax
  \providecommand{\doi}[1]{doi: #1}\else
  \providecommand{\doi}{doi: \begingroup \urlstyle{rm}\Url}\fi

\bibitem[Bashivan et~al.(2016)Bashivan, Rish, Yeasin, and Codella]{bashivan}
Pouya Bashivan, Irina Rish, Mohammed Yeasin, and Noel Codella.
\newblock Learning representations from eeg with deep recurrent convolutional
  neural networks.
\newblock In \emph{Proceedings of the International Conference on Learning
  Representations}, 2016.

\bibitem[Chollet(2015)]{chollet2015keras}
Fran\c{c}ois Chollet.
\newblock Keras.
\newblock \url{https://github.com/fchollet/keras}, 2015.

\bibitem[CP(2010)]{clinical_seizure}
Panayiotopoulos CP.
\newblock A clinical guide to epileptic syndromes and their treatment.
\newblock chapter~6. Springer, 2010.

\bibitem[Erhan et~al.(2010)Erhan, Bengio, Courville, Manzagol, Vincent, and
  Bengio]{unsupervised}
Dumitru Erhan, Yoshua Bengio, Aaron Courville, Pierre-Antoine Manzagol, Pascal
  Vincent, and Samy Bengio.
\newblock Why does unsupervised pre-training help deep learning?
\newblock 2010.

\bibitem[Fotiadis(2016)]{chbmit_pat}
Dimitrios~I. Fotiadis.
\newblock \emph{Handbook of Research on Trends in the Diagnosis and Treatment
  of Chronic Conditions}.
\newblock IGI Global book series advances in Medical Diagnosis, Treatment, and
  Care, 2016.

\bibitem[Furbass et~al.(2014)Furbass, P, M, H, AM, G, L, E, AJ, C, and
  T]{multi}
F~Furbass, Ossenblok P, Hartmann M, Perko H, Skupch AM, Lindinger G, Elezi L,
  Patarai E, Colon AJ, Baumgartner C, and Kluge T.
\newblock Prospective multi-center study of an automatic online seizure
  detection system for epilepsy monitoring units.
\newblock \emph{Clinical Neurophysiology}, 2014.

\bibitem[Goodfellow et~al.(2016)Goodfellow, Bengio, and
  Courville]{Goodfellow-et-al-2016-Book}
Ian Goodfellow, Yoshua Bengio, and Aaron Courville.
\newblock Deep learning.
\newblock Book in preparation for MIT Press, 2016.
\newblock URL \url{http://www.deeplearningbook.org}.

\bibitem[Gotman(1999)]{gotman1}
Jean Gotman.
\newblock Automatic detection of seizures and spikes.
\newblock \emph{Journal of Clinical Neurophysiology}, 16:\penalty0 130--140,
  1999.

\bibitem[Graves and Schmidhuber(2005)]{graves2005framewise}
Alex Graves and J{\"u}rgen Schmidhuber.
\newblock Framewise phoneme classification with bidirectional lstm and other
  neural network architectures.
\newblock \emph{Neural Networks}, 18\penalty0 (5):\penalty0 602--610, 2005.

\bibitem[Hensman and Masko(2015)]{imbalance}
Paulina Hensman and David Masko.
\newblock The impact of imbalanced training data for convolutional neural
  networks, 2015.

\bibitem[Hochreiter and Schmidhuber(1997)]{Hochreiter:1997:LSM:1246443.1246450}
Sepp Hochreiter and J\"{u}rgen Schmidhuber.
\newblock Long short-term memory.
\newblock \emph{Neural Comput.}, 9\penalty0 (8):\penalty0 1735--1780, November
  1997.
\newblock ISSN 0899-7667.
\newblock \doi{10.1162/neco.1997.9.8.1735}.
\newblock URL \url{http://dx.doi.org/10.1162/neco.1997.9.8.1735}.

\bibitem[KM et~al.(2012)KM, DS, RT, JH, MCK, KA, JP, JJ, and
  JC]{clinical_bench}
Kelly KM, Shiau DS, Kern RT, Chien JH, Yang MCK, Yandora KA, Valeriano JP,
  Halford JJ, and Sackellares JC.
\newblock Assessment of a scalp eeg-based automated seizure detection system.
\newblock \emph{Clinical Neurophysiology}, 121, 2012.

\bibitem[Krizhevsky et~al.(2012)Krizhevsky, Sutskever, and
  Hinton]{NIPS2012_4824}
Alex Krizhevsky, Ilya Sutskever, and Geoffrey~E. Hinton.
\newblock Imagenet classification with deep convolutional neural networks.
\newblock In F.~Pereira, C.~J.~C. Burges, L.~Bottou, and K.~Q. Weinberger,
  editors, \emph{Advances in Neural Information Processing Systems 25}, pages
  1097--1105. Curran Associates, Inc., 2012.
\newblock URL
  \url{http://papers.nips.cc/paper/4824-imagenet-classification-with-deep-convolutional-neural-networks.pdf}.

\bibitem[LeCun and Bengio(1995)]{lecun-bengio-95a}
Yann. LeCun and Yoshua. Bengio.
\newblock Convolutional networks for images, speech, and time-series.
\newblock In M.~A. Arbib, editor, \emph{The Handbook of Brain Theory and Neural
  Networks}. MIT Press, 1995.

\bibitem[Megiddo et~al.(2016)Megiddo, A, D, T, A, and R]{epilepsy_world}
Megiddo, Colson A, Chisholm D, Dua T, Nandi A, and Laxminarayan R.
\newblock Health and economic benefits of public financing of epilepsy
  treatment in india: An agent-based simulation model.
\newblock \emph{Epilepsia Official Journal of the International League Against
  Epilepsy}, 13294, 2016.

\bibitem[Mormann et~al.(2007)Mormann, Andrzejak, Elger, and
  Lehnertz]{mormann07}
Florian Mormann, Ralph~G. Andrzejak, Christian~E. Elger, and Klaus Lehnertz.
\newblock Seizure prediction: the long and winding road.
\newblock \emph{Brain}, 130:\penalty0 314--333, 2007.

\bibitem[Shoeb(2009)]{shoeb}
Ali Shoeb.
\newblock \emph{Application of Machine Learning to Epileptic Seizure Onset
  Detection and Treatment}.
\newblock PhD thesis, Massachusetts Institute of Technology, 2009.

\bibitem[SJ(2005)]{offline_eeg}
Smith SJ.
\newblock Eeg in the diagnosis, classification, and management of patients with
  epilepsy.
\newblock \emph{Journal of Neurology, Neurosurgery and Psychiatry}, 2005.

\bibitem[Snyder and Parr(1987)]{polar_proj}
Snyder and Jone Parr.
\newblock Map projections-a working manual, 1987.

\bibitem[Tzallas et~al.(2012)Tzallas, Tsipouras, Tsalikakis, Karvounis,
  Astrakas, Konitsiotis, and Tzaphlidou]{review_seizure}
Alexandros~T. Tzallas, Markos~G. Tsipouras, Dimitrios~G. Tsalikakis,
  Evaggelos~C. Karvounis, Loukas Astrakas, Spiros Konitsiotis, and Margaret
  Tzaphlidou.
\newblock Automated epileptic seizure detection methods: A review study.
\newblock \emph{Epilepsy - Histological, Electroencephalographic and
  Psychological Aspects}, 2012.

\bibitem[Wilson et~al.(2004)Wilson, Scheuer, Emerson, and Gabor]{reveal_cross}
Scott~B. Wilson, Mark~L. Scheuer, Ronald~G. Emerson, and Andrew~J. Gabor.
\newblock Seizure detection: evaluation of the reveal algorithm.
\newblock \emph{Clinical Neurophysiology}, 115:\penalty0 2280--2291, 2004.

\bibitem[Yao et~al.(2007)Yao, Rosasco, and Caponnetto]{early_stopping}
Yuan Yao, Lorenzo Rosasco, and Andrea Caponnetto.
\newblock On early stopping in gradient descent learning.
\newblock \emph{Constructive Approximation}, 26, 2007.

\bibitem[Zhou et~al.(2002)Zhou, Wu, and Tang]{zhou2002ensembling}
Zhi-Hua Zhou, Jianxin Wu, and Wei Tang.
\newblock Ensembling neural networks: many could be better than all.
\newblock \emph{Artificial intelligence}, 137\penalty0 (1):\penalty0 239--263,
  2002.

\end{thebibliography}

\appendix


\end{document}